\documentclass{article}

\PassOptionsToPackage{numbers, compress}{natbib}

 \usepackage[preprint]{neurips_2026}

\usepackage[utf8]{inputenc} %
\usepackage[T1]{fontenc}    %
\usepackage{url}            %
\usepackage{booktabs}       %
\usepackage{amsfonts}       %
\usepackage{nicefrac}       %
\usepackage{microtype}      %
\usepackage{xcolor}         %

\usepackage{graphicx}
\usepackage{subcaption}
\usepackage{algorithm}
\usepackage{algpseudocode}
\usepackage{amsmath}
\usepackage{wrapfig}
\usepackage{enumitem}

\usepackage{xurl}
\usepackage{xcolor}

\definecolor{vlayellow}{RGB}{210,125,0}
\definecolor{chunkcream}{RGB}{255,190,20}
\definecolor{wmgreen}{RGB}{95,155,85}
\definecolor{encoderblue}{RGB}{90,150,190}
\definecolor{dreamgreen}{RGB}{190,225,175}
\definecolor{stateblue}{RGB}{175,210,225}

\definecolor{bluegray}{rgb}{0.4, 0.6, 0.8}
\definecolor{lightcarminepink}{rgb}{0.9, 0.4, 0.38}
\usepackage[pagebackref,breaklinks,colorlinks,citecolor=bluegray,linkcolor=lightcarminepink]{hyperref}
\newcommand{\OURS}{\texttt{DREAM-Chunk$\,$}}

\usepackage{multirow}

\makeatletter
\usepackage{xspace}
\def\@onedot{\ifx\@let@token.\else.\null\fi\xspace}
\DeclareRobustCommand\onedot{\futurelet\@let@token\@onedot}

\newcommand{\figref}[1]{Fig\onedot~\ref{#1}}

\newcommand{\secref}[1]{Sec\onedot~\ref{#1}}
\newcommand{\tabref}[1]{Tab\onedot~\ref{#1}}

\newcommand{\algoref}[1]{Alg\onedot~\ref{#1}} %

\usepackage[aboveskip=1pt]{subcaption}

\title{DREAM-Chunk: Reactive Action Chunking\\ with Latent World Model
}

\author{%
  \textbf{Wenxi Chen$^{1}$} \quad
  \textbf{Kaidi Zhang$^{1,\dagger}$} \thanks{Kaidi Zhang and Chi Lin made equal contributions to the experimental work.}  \quad
  \textbf{Chi Lin$^{1,\dagger}$} \quad
  \textbf{Zhiyuan Zhang$^{1}$} \quad
  \textbf{Yu She$^{1}$} \\
  \textbf{Yuejiang Liu$^{2}$} \quad
  \textbf{Raymond A. Yeh$^{1}$} \quad
  \textbf{Shaoshuai Mou$^{1}$} \quad
  \textbf{Yan Gu$^{1}$} \\
  $^{1}$Purdue University \quad
  $^{2}$Stanford University \\
}

\begin{document}

\maketitle

 \begin{abstract}
Action chunking has become a common interface for vision-language-action (VLA) models, enabling low-frequency policy inference to drive high-frequency robot execution. However, once a chunk is committed, its open-loop execution can be brittle under stochastic dynamics, hardware execution errors, and partially observed state. We propose \OURS, a test-time scaling method that augments chunking-based policies with a lightweight latent world model without requiring additional policy fine-tuning. At test time, \OURS samples multiple candidate action chunks, rolls out their predicted latent futures, and selects actions from the chunk whose ``dreamed'' state best matches the observed rollout. In this way, \OURS uses additional test-time computation to cover multiple plausible stochastic rollouts and improve reactivity during long-horizon chunk execution. On the Kinetix benchmark, \OURS improves robustness under increasing action noise and benefits from larger candidate sample sizes, especially when demonstrations contain corrective behaviors. We conduct empirical validation on four manipulation tasks across two robot platforms and two VLA policies for various stochasticity sources. Across simulation and hardware experiments, \OURS improves action-chunking policies in stochastic dynamics. Project page: \url{https://wenxichen2746.github.io/DREAM-Chunk/}.
\end{abstract}

\section{Introduction}

\begin{wrapfigure}[17]{r}{0.48\textwidth}
  \centering
  \vspace{-0.5cm}
    \includegraphics[
    width=0.47\textwidth,
    clip
  ]{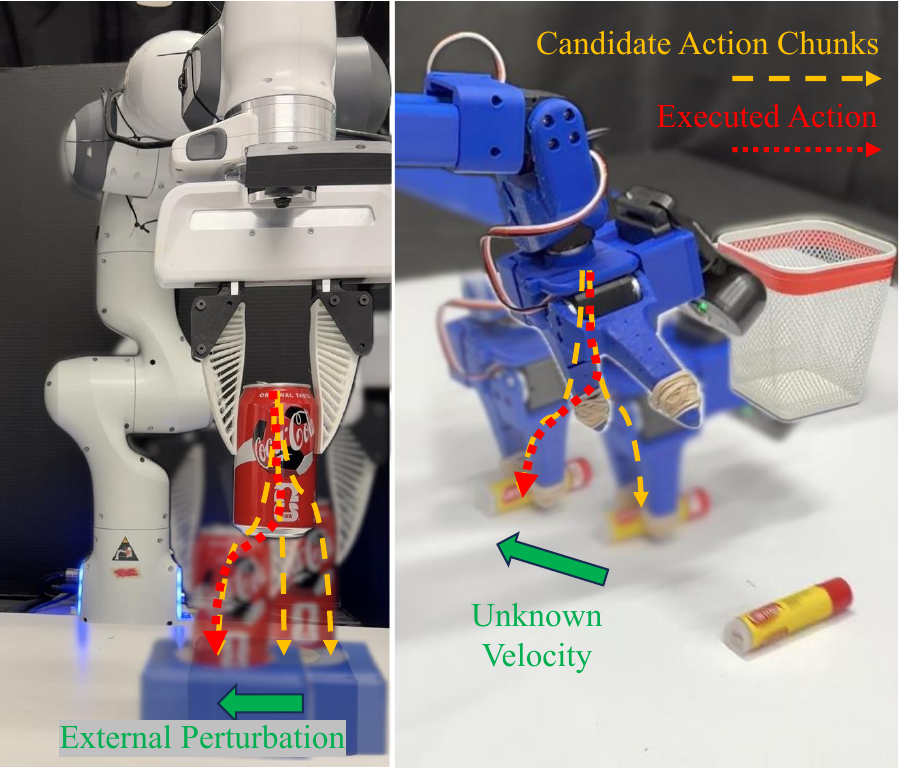}
  \vspace{-0.1cm}
  \caption{
    Illustration of chunk switch to handle external perturbation and stochastic dynamics. 
    }
  \label{fig:hardware_demo}
  \vspace{-10pt}
\end{wrapfigure}

Vision-language-action (VLA) models aim to bring foundation-model capabilities into physical control by learning language-conditioned visuomotor policies from large-scale robot demonstration data~\cite{zitkovich2023rt,team2025gemini,black2025pi}. In parallel, world models provide a complementary route toward embodied intelligence by learning predictive structure from trajectory data, enabling models to reason about how future states evolve under actions. Recent VLA studies have begun to connect these two directions, for example, by incorporating video prediction or world-action modeling objectives to improve spatiotemporal understanding~\cite{ye2026gigaworld,cen2025worldvla,ni2025swiftvla}. The recent $\pi_{0.7}$ system from Physical Intelligence~\cite{physicalintelligence2026pi07} also reflects this trend, using a video-generation world model to provide context for VLA inference. These developments suggest a modular path toward practically deployable embodied intelligence~\cite{kim2026modular}, integrating advances across complementary areas of robotics research, including VLA and world model studies~\cite{hou2026world}.

A widely used approach in learning from demonstration is action chunking~\cite{zhao2023learning}, where a policy predicts a short sequence of future actions rather than a single-step action. By modeling temporally coherent action segments, action chunking has been effective for imitation learning and generative visuomotor policy learning~\cite{chi2025diffusion,pan2025much}. Recent VLAs have increasingly adopted this chunked action interface, in part because it allows low-frequency policy inference to drive higher-frequency robot execution. Representative VLA models either use generative ``action experts'' to directly sample action chunks, as in $\pi_{0.5}$~\cite{black2025pi,intelligence2025pi}, Octo~\cite{team2024octo}, and SmolVLA~\cite{shukor2025smolvla}, or predict action tokens that are decoded into short action chunks, as in RT-2~\cite{zitkovich2023rt}, OpenVLA~\cite{kim2024openvla}, and FAST~\cite{pertsch2025fast,intelligence2025pi,physicalintelligence2026pi07}.

This chunked interface makes large VLAs more practical for real-time deployment. The interface has been shown to help handle inference latency through asynchronous execution~\cite{shukor2025smolvla,tang2025vlash} while preserving mode consistency when learning from diverse demonstrations~\cite{chi2025diffusion,zhao2023learning}. However, these benefits come with an important limitation: once a chunk is committed, the robot executes it largely open-loop, causing errors to accumulate during execution~\cite{laskey2017dart}. \citet{liu2025bidirectional} analyzes that long-horizon execution can provide an information advantage when the policy's implicit internal model correctly predicts unobserved future states, but this advantage can be substantially weakened under stochastic dynamics.

Recent efforts to improve VLAs' reactivity include repeatedly replanning new action chunks while conditioning on future states~\cite{tang2025vlash}, training lightweight VLAs for faster inference~\cite{xie2026dynamicvla}, adjusting execution horizons adaptively~\cite{so2025improving,liang2026adaptive}, and training residual modules to correct generated chunks~\cite{xu2026rlt,sendai2025leave}. However, existing approaches still face important limitations. Frequent replanning can be computationally expensive and may waste the long-horizon plans already predicted by the policy, while improving reactivity through lightweight architectural modifications may limit the potential of scaling to larger models and broader training tasks.

Motivated by the emerging connection between VLA policies and world models, we ask whether lightweight prediction can improve long-horizon chunk execution at test time without modifying the base policy. We propose \textbf{Dreamed-state REactive Action Matching for Action Chunking (\OURS)}, which keeps the base policy fixed, samples multiple candidate chunks, and predicts their latent futures with a lightweight world model. During execution, \OURS matches the observed state to phase-aligned predicted rollouts and selects the action from the most consistent candidate chunk, enabling reactive switching when stochastic dynamics move the robot away from the nominal rollout. \figref{fig:hardware_demo} illustrates this process: yellow dashed lines denote candidate rollouts, and the green arrow denotes stochastic effects such as external perturbations or varying object velocities. The chunk horizon and mode differences are exaggerated for clarity.

{\bf Our main contributions are threefold:}
{\bf (i)} We propose \OURS, a policy-agnostic test-time scaling framework that improves the reactivity of action-chunking policies under stochastic dynamics without modifying or fine-tuning the base VLA.
{\bf (ii)} In controlled Kinetix simulations~\cite{matthews2025kinetix}, \OURS outperforms representative test-time chunking baselines under high stochasticity, and our analysis studies the effects of corrective expert demonstrations and latent representations from R2-Dreamer~\cite{morihira2026rdreamer}, LeWorldModel~\cite{maes2026leworldmodel}, and EB-JEPA~\cite{terver2026lightweight}.
{\bf (iii)} We validate \OURS on four real-world manipulation tasks across SO-101 and Franka Panda using SmolVLA and $\pi_{0.5}$, covering hardware execution errors, partial observability, and external perturbations; across these settings, \OURS improves success rates, including increasing open-loop $\pi_{0.5}$ from 10\% to 65\% on a precise insertion task under external perturbation.

\section{Related Works}
{\bf\noindent Test-time scaling for action chunking.}
We use test-time scaling broadly to refer to methods that allocate additional computation or modify the sampling procedure at inference time to improve action-chunking policies.
Bidirectional Decoding (BID)~\cite{liu2025bidirectional} samples multiple candidate chunks and selects one that is consistent with the previous trajectory while avoiding actions similar to those from a poor policy.
Real-Time Action Chunking (RTC)~\cite{black2026realtime} guides the sampling of a new action chunk via inpainting so that the new chunk aligns with the previously executed chunk, which requires backpropagation at test time.
Self-Guidance~\cite{so2025improving} applies classifier-free-style guidance during action sampling, using two forward passes of both current and previous observations to push actions away from distributions induced by prior observations. They also propose adaptive chunking to balance reactivity and consistency, which repeatedly infers new chunks and switches execution when the new action differs sufficiently from the current chunk. Recent methods use uncertainty across batched chunk samples to adapt execution: \citet{liang2026adaptive} adjust the execution horizon based on action entropy, while HiPolicy~\cite{zhang2026hipolicy} samples coarse-to-fine chunks at multiple frequencies and uses entropy to determine the execution frequency. Beyond chunk-specific mechanisms, RoboMonkey~\cite{kwok2025robomonkey} uses test-time action sampling and a fine-tuned VLM verifier to select among candidate actions. 
In contrast, \OURS treats chunk execution itself as the test-time decision problem: it samples multiple chunks once and selects among their phase-aligned actions using new observations during execution.

{\bf\noindent World models} have emerged as another major path toward generalist intelligence, spanning diverse efforts from video generation~\cite{alonso2024diffusion} to model-based reinforcement learning~\cite{hafner2019dream}. One line of work aims to learn foundation-scale models that predict future observations and actions, where visual reconstruction provides an auxiliary training signal for spatiotemporal understanding and enables scaling from large-scale video data~\cite{ye2026gigaworld,cen2025worldvla,ni2025swiftvla}. Another line instead learns latent dynamics models without requiring a visual decoder, using the learned latent transition structure for policy learning~\cite{hafner2019dream,morihira2026rdreamer} or online action planning~\cite{maes2026leworldmodel,terver2026lightweight}. Efforts have been made to leverage latent world models to improve the performance of VLA models. For instance, FOREWARN~\cite{wu2025forewarn} uses a latent world model to predict candidate action outcomes and lets a latent-aligned VLM choose the best action plan based on high-level user intent and semantic constraints.

\section{Preliminary}
\label{sec:preliminary}
We briefly review the necessary background to establish a common notation. %

{\bf\noindent Action chunking policy.} 
We consider an action-chunking policy parameterized as a generative model $\pi(A_t \mid o_t)$ taking observation $o_t$ as input, where each action chunk $A_t = \{a_t, a_{t+1}, \dots, a_{t+L-1}\}$ is a sequence of actions with chunk length $L$. The policy is trained from an offline dataset of expert demonstrations $\{(o_t, a_t)\}$ by minimizing the discrepancy between generated action chunks and action segments sliced from the demonstration trajectories. At test time, the policy typically executes only part of each predicted chunk before replanning. We denote this effective number of executed actions as the execution horizon $H$.

\begin{figure}[t]
  \vspace{-3pt}
  \centering
  \includegraphics[width=.91\textwidth]{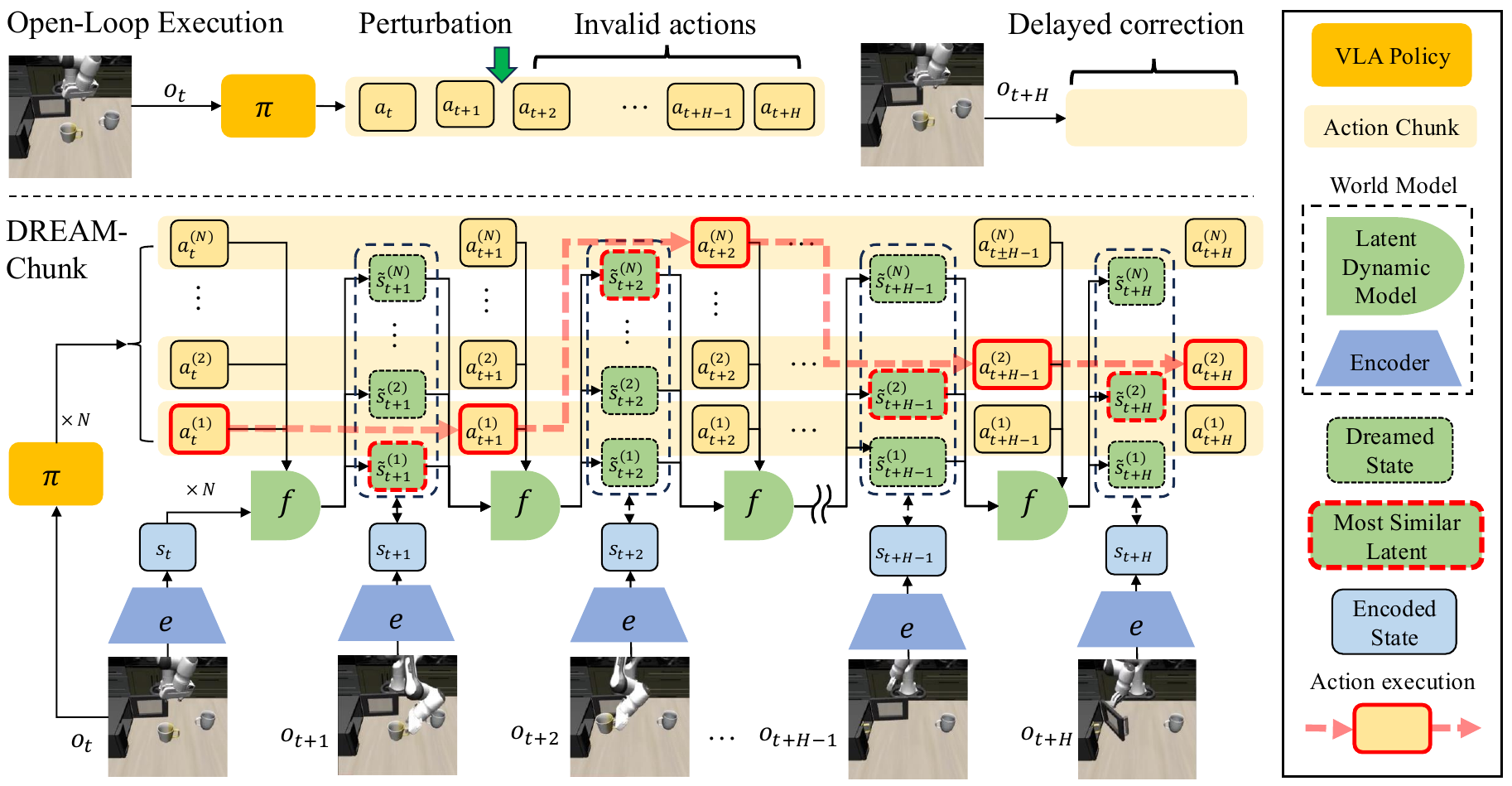}
  \vspace{-5pt}
  \caption{
Naive open-loop action chunk execution cannot correct actions until the next inference step. \OURS samples batched {\color{chunkcream}action chunks} from {\color{vlayellow} VLA} and switches among them using world-model predictions. The latent world model, which is much more lightweight than VLA, predicts future rollouts from the current observation: each new observation is encoded by the {\color{encoderblue}encoder} and compared with predictions from the {\color{wmgreen}latent dynamics model}. The executed action, indicated by the {\color{red}dashed red arrow}, is selected from the chunk whose dreamed state best matches the real-world rollout. This allows sampled chunks to provide alternative rollouts that keep actions valid under stochasticity.
}
  \label{fig: dreamchunkdiagram}
  \vspace{-0.49cm}
\end{figure}

{\bf\noindent Reactive long-horizon control.}
Since the whole chunk is generated from the observation available at inference time, later actions are executed based on increasingly outdated state information. To formalize this mismatch, let $A_t=\{a_t,\ldots,a_{t+H-1}\}\sim\pi(\cdot\mid o_t)$ denote an action chunk sampled at time $t$, and let $A_t[\tau]$ be its $\tau$-th action. Ideally, the action executed at phase $\tau$ should remain aligned with the decision that the same policy would make if it were re-inferred from the realized observation $o_{t+\tau}$. For an executed action $a_{t+\tau}$, we define this alignment as
\begin{equation}
    \mathrm{Align}(a_{t+\tau}, o_{t+\tau})
    =
    \mathbb{E}_{\hat{A}_{t+\tau}\sim\pi(\cdot\mid o_{t+\tau})}
    \left[
        \exp\left(
        -\left\|a_{t+\tau}-\hat{A}_{t+\tau}[0]\right\|_2^2
        \right)
    \right],
\end{equation}
where $\hat{A}_{t+\tau}[0]$ is the first action of a chunk inferred from the current observation $o_{t+\tau}$.

{\bf\noindent Objective.}
We use the expected alignment as a measure of reactivity for a chunk execution strategy $\eta$, which selects the executed action $a^{\eta}_{t+\tau}$ from available action chunks during rollout. Here, $P_{\eta}(o_{t+1:t+H}\mid o_t)$ denotes the rollout distribution induced by the stochastic dynamics $o_{t+\tau+1}\sim P(\cdot\mid o_{t+\tau},a^{\eta}_{t+\tau})$. We define the reactivity objective
\begin{equation}
    J_{\mathrm{react}}(\eta)
    =
    \mathbb{E}_{o_{t+1:t+H}\sim P_{\eta}(\cdot\mid o_t)}
    \left[
        \sum_{\tau=0}^{H-1}
        \mathrm{Align}(a^{\eta}_{t+\tau}, o_{t+\tau})
    \right].
\end{equation}
A more reactive execution strategy achieves a larger $J_{\mathrm{react}}(\eta)$ by keeping executed actions aligned with what the policy would choose under realized observations. Stochastic dynamics can move the rollout away from the nominal trajectory of the original chunk, causing later open-loop actions to become misaligned with the current policy decision. We denote the per-step probability of such deviation as $p_{\mathrm{dyn}}$. Directly evaluating $J_{\mathrm{react}}(\eta)$ would require repeatedly re-inferring the VLA at each realized observation; \OURS instead uses latent matching as a lightweight surrogate, selecting candidate chunks whose dreamed latent states best match the realized rollout.

\section{Method}
\label{sec:method}
\OURS augments a chunking-based VLA policy with an auxiliary world model and defines an execution strategy over candidate chunks sampled from the fixed policy $\pi$. As shown in~\figref{fig: dreamchunkdiagram}, the method maintains a set of candidate action chunks, rolls out their latent consequences with the world model, and reactively selects which chunk to follow based on latent-state compatibility. 

{\bf\noindent Auxiliary world model.}
\OURS uses an auxiliary world model consisting of an encoder and a latent dynamics model. The encoder maps high-dimensional observations, such as images and robot joint measurements, into a latent state space, $s_t = e(o_t)$. The latent dynamics model then predicts the next latent state from the current latent state and action, $\tilde{s}_{t+1} = f(s_t, a_t)$. Since the world model is trained on the same offline expert demonstrations as the action-chunking policy, we expect it to capture latent transition structure that is broadly aligned with the policy's implicit internal model~\cite{liu2025bidirectional}.

{\bf\noindent Intuition.}
\OURS samples multiple plausible short-horizon plans from the current observation and uses an auxiliary world model to predict their latent rollouts. During execution, it selects the chunk whose predicted latent state best matches the observed state, allowing the robot to switch away from a rollout that has become inconsistent under stochastic dynamics. \OURS therefore does not invent new recovery behaviors at test time; instead, it exposes and selects corrective behaviors already present in the policy's sampled action distribution. We further analyze when \OURS can outperform high-frequency short-horizon control in Appendix~\ref{app:reactivity_coverage}.

\begin{figure}[t]
\begin{algorithm}[H]
\caption{\OURS Execution under Asynchronous Inference}
\label{alg:reactive_chunk_selection}
\begin{algorithmic}[1]
\Require chunking policy $\pi$, encoder $e$, latent dynamics model $f$
\Require execution horizon $H$, overlap step $d$, number of candidate chunks $N$

\State $\mathcal{A}=\{A^{(j)}\}_{j=1}^N \gets \pi(o_t)$
\State $s_t \gets e(o_t)$
\State $\tilde{S}_t \gets \mathrm{Repeat}(s_t,N)$ \Comment{$\tilde{S}_t\in\mathbb{R}^{N\times d_z}$ for batched rollout}

\While{not terminated}
    \For{$\tau=0,\dots,H-1$}
        \State $s_{t+\tau}\gets e(o_{t+\tau})$

        \State $i\gets \arg\min_{j\in\{1,\dots,N\}}
        \left\|s_{t+\tau}-\tilde{S}_{t+\tau}[j]\right\|_2$

        \State $a_{t+\tau}\gets A^{(i)}_\tau$; execute $a_{t+\tau}$
        \If{$\tau=H-1-d$}
            \State $S\gets \{s_{t+\tau}\}_{j=1}^N$; \quad $\mathcal{A}^{\mathrm{next}}\gets\pi(o_{t+\tau})$ asynchronously
        \EndIf
        
        \State $\tilde{S}_{t+\tau+1}\gets 
        f\!\left(\tilde{S}_{t+\tau}, \{A^{(j)}_{\tau}\}_{j=1}^N\right)$
    \EndFor
    \State $\mathcal{A}\gets\mathcal{A}^{\mathrm{next}}$; \quad $\tilde{S}\gets{S}$; \quad $i\gets 1$
\EndWhile
\end{algorithmic}
\end{algorithm}
\vspace{-0.8cm}
\end{figure}

\subsection{Batched Action Chunks Inference and Parallel Latent Rollouts}
\label{subsec:test_time_scaling}

At each replanning step $t$, \OURS samples $N$ candidate action chunks from the chunking policy,
\begin{equation}
    \mathcal{A}=\{A^{(j)}\}_{j=1}^N \gets \pi(o_t),
    \qquad
    A^{(j)} = a^{(j)}_{t:t+H-1}.
\end{equation}
The current observation $o_t$ is encoded as $s_t=e(o_t)$, and this latent state is duplicated across the candidate dimension to initialize a batched latent tensor,
\begin{equation}
    \tilde{S}_t = \mathrm{Repeat}(s_t,N),
    \qquad
    \tilde{S}_t \in \mathbb{R}^{N\times d_z}.
\end{equation}

Starting from $\tilde{S}_t$, \OURS rolls out the latent consequences of all candidate chunks in parallel. At each phase $\tau=0,\dots,H-1$, the batched latent dynamics model updates
\begin{equation}
    \tilde{S}_{t+\tau+1}
    =
    f\!\left(
        \tilde{S}_{t+\tau},
        \{A^{(j)}_\tau\}_{j=1}^N
    \right),
\end{equation}
where $\tilde{S}_{t+\tau}[j]$ denotes the dreamed latent state for candidate chunk $j$ at execution phase $\tau$ and $A^{(j)}_\tau$ is the $\tau$-th action in candidate chunk $A^{(j)}$. This produces a phase-aligned latent rollout for each candidate chunk. Importantly, the dreamed states are not reset using future observations during the rollout; they represent the expected latent evolution induced by each sampled action chunk.

This procedure acts as a form of test-time scaling: additional inference-time computation is used to sample more candidate chunks and predict more possible latent futures. Instead of committing to a single open-loop chunk, \OURS maintains a batched set of imagined rollouts that can be queried online as new observations arrive.

\subsection{Execution via Reactive Latent Matching}
\label{subsec:reactive_latent_matching}

During execution, the robot receives a new observation $o_{t+\tau}$ at each phase $\tau \in \{0,\dots,H-1\}$ and encodes it as
$
    s_{t+\tau}=e(o_{t+\tau})
$.
\OURS compares this observed latent state with the phase-aligned dreamed states from the batched rollout and selects the candidate chunk whose prediction best matches the realized state:
\begin{equation}
    i^\star
    =
    \arg\min_{j\in\{1,\dots,N\}}
    \left\|
    s_{t+\tau} - \tilde{S}_{t+\tau}[j]
    \right\|_2 .
\end{equation}
The robot then executes the phase-aligned action $A^{(i^\star)}_\tau$ from the selected chunk.

This rule allows \OURS to switch among candidate chunks when the observed trajectory deviates from the predicted rollout of the currently followed chunk. Because the matching with the observed latent state is performed only against dreamed states at the same execution phase, the switch remains temporally aligned with the action sequence rather than jumping to an earlier or later stage of another rollout. \OURS can also be incorporated with asynchronous inference by storing the true latent state at policy inference, as stated in~\algoref{alg:reactive_chunk_selection}.

\subsection{Auxiliary World Model Formulation}
\label{subsec:world_model_learning}
 The latent state representation could be deterministic, following the line of works on JEPA~\cite{lecun2022path,maes2026leworldmodel}, or stochastic ($s=(h,z)$, where $h$ is deterministic, and $z$ is stochastic) as formulated in the Recurrent State-Space Model (RSSM) adopted in the Dreamer series~\cite{hafner2023mastering, morihira2026rdreamer}. Our experiment will show that the \OURS method can be adapted to both formulations as shown in our ablation study in~\figref{fig:kinetix_wm_ablations}. Further, we want to highlight that this latent dynamics model is substantially lighter than VLA models and does not introduce significant computation overhead. For example, a JEPA world model can only have 15 million parameters, while a small-scale VLA like SmolVLA has around 450 million parameters, and larger models like $\pi_{0.5}$ can have more than 2 billion parameters. In our hardware experiments, VLA inference takes more than 100 ms, and world model encoding and prediction take less than 10 ms, as listed in~\tabref{tab:inference_latency}.

\begin{figure}[t]
  \centering
  \begin{subfigure}[h]{0.63\textwidth}
    \centering
    \includegraphics[width=\textwidth]{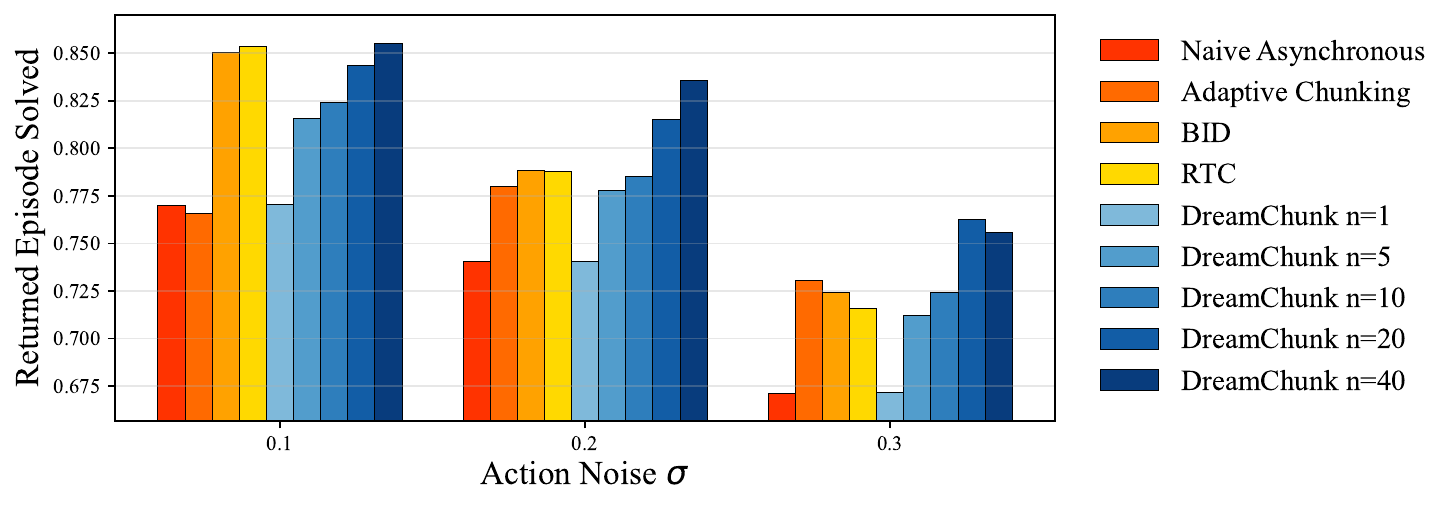}
    \caption{Returned episode solved.}
    \label{fig:kinetix_actionnoise}
  \end{subfigure}
  \hfill
  \begin{subfigure}[h]{0.36\textwidth}
    \centering
    \includegraphics[width=\textwidth]{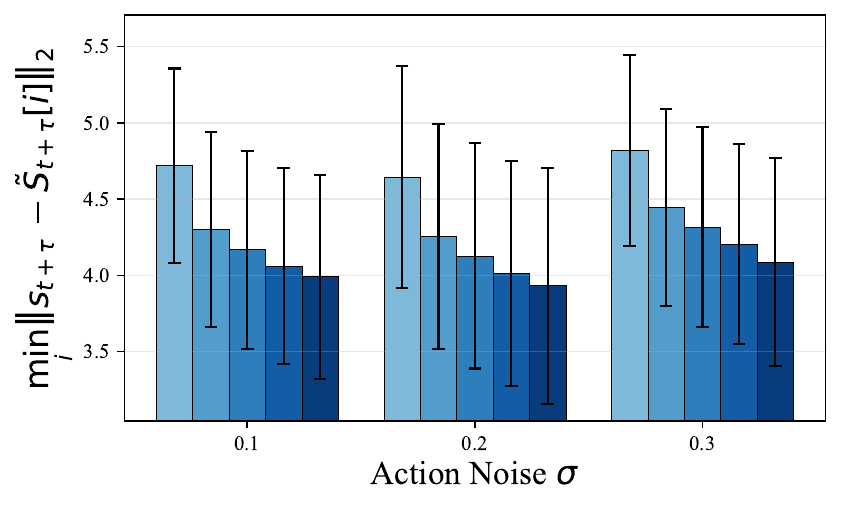}
    \caption{Similarity between predicted states.}
    \label{fig:kinetix_actionnoise_sim_coverage}
  \end{subfigure}
  \vspace{-5pt}
  \caption{Kinetix simulation result on performance and latent similarity. Reported values are averaged over 12 environments.}
  \label{fig:kinetix_actionnoise_both}
  \vspace{-12pt}
\end{figure}

\section{Experiment}
We evaluate \OURS in both controlled simulation and real-world hardware settings. In~\secref{sec:kinetix_simulation}, we use the Kinetix benchmark to study how \OURS behaves under controlled action noise, including the effects of sample count, corrective behaviors in demonstrations, and latent world model design. In~\secref{sec:hardware_experiment}, we deploy \OURS on real robot manipulation tasks to test whether the same mechanism improves robustness under real-world stochasticity like hardware execution imperfections, partial observability, and external perturbations.

\subsection{Kinetix Simulation}
\label{sec:kinetix_simulation}
\begin{wrapfigure}[18]{r}{0.42\textwidth}
  \centering
  \vspace{-27pt}
  \includegraphics[width=0.40\textwidth]{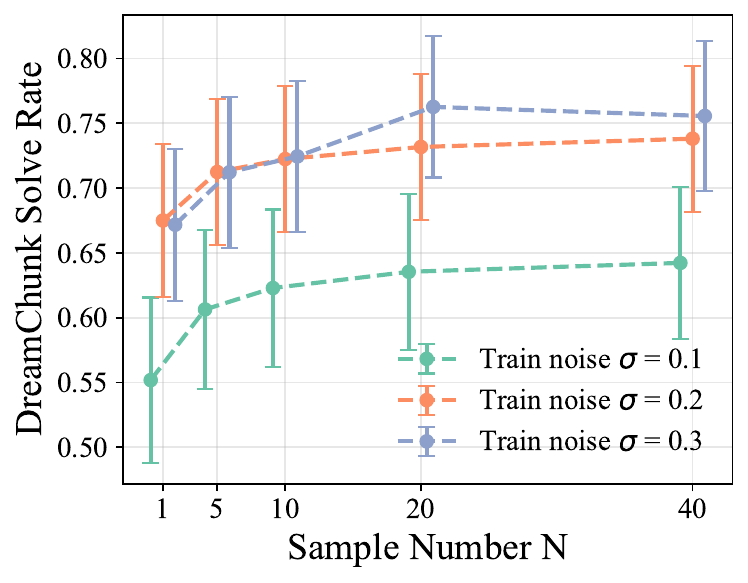}
  \vspace{-0.2cm}
  \caption{
    The figure reports the solve rate of DreamChunk under action noise $0.3$ with different sample size $N$, where the behavior cloning policy is trained on demonstrations from experts trained with different action-noise levels. 
    }
  \label{fig:kinetix_expert_train_noise}
  \vspace{-5pt}
\end{wrapfigure}

Our simulation experiments are designed to answer three questions:
(1) Does increasing the number of sampled chunks improve \OURS's performance under stochastic dynamics?
(2) What conditions are necessary for test-time scaling with \OURS to be effective?
(3) How does the learned latent representation of the auxiliary world model affect performance?

Kinetix~\cite{matthews2025kinetix} is a 2D agentic simulation benchmark spanning locomotion, manipulation, and Atari-like control tasks. Following the RTC evaluation setting~\cite{black2026realtime}, we train a flow-matching action-chunking policy by imitation learning from RL expert demonstrations collected with different random seeds. We introduce environmental stochasticity by adding Gaussian noise to the executed action, $a_{\text{exec}} = a + \epsilon$, where $\epsilon \sim \mathcal{N}(0, \sigma^2 I)$ and $\sigma$ controls the noise level.

{\bf\noindent Implementation detail.} We compare \OURS against naive asynchronous execution~\cite{shukor2025smolvla,tang2025vlash}, RTC~\cite{black2026realtime}, BID~\cite{liu2025bidirectional}, and Adaptive Chunking (AC)~\cite{so2025improving}. For BID and RTC, we follow the settings provided in~\cite{black2026realtime}. BID uses 10 sampled action chunks and uses an early checkpoint as a weak policy. We only display the performance of the RTC variant with soft-masking as championed by the original paper. We implement AC following~\cite{so2025improving}, using cosine similarity as the switching metric with a threshold of $0.97$. Unless otherwise specified, reported \OURS performance uses an RSSM-based world model trained with one-step prediction loss and Barlow Twins regularization~\cite{morihira2026rdreamer}.

{\bf\noindent Performance under stochasticity.}
As shown in~\figref{fig:kinetix_actionnoise}, \OURS gains larger improvements as the number of sampled chunks increases. This comparison highlights the benefit of improving \emph{within-chunk reactivity}. In contrast, BID and RTC mainly target \emph{cross-chunk consistency} by improving the transition between independently inferred chunks. Under low stochasticity, both directions provide similar gains over naive asynchronous execution. However, as the variance of the action noise increases, \OURS achieves a larger advantage, suggesting that within-chunk reactivity becomes more important when the realized trajectory deviates from the nominal open-loop rollout. AC also aims to improve reactivity by adaptively changing the execution horizon, but its performance is often close to the naive one-step replanning baseline on dynamic tasks, suggesting that it tends to replan frequently rather than preserve longer chunk execution. We provide additional results across execution horizons in Appendix~\figref{fig:kinetix_executionhorizon}, where \OURS shows particular strength under mid- to long-horizon execution.

To test whether more candidate chunks provide closer predicted futures to the states actually reached under stochastic execution, we measure the average nearest-neighbor distance between the encoded latent state reached during execution and the dreamed latent states predicted by candidate chunks. As shown in \figref{fig:kinetix_actionnoise_sim_coverage}, increasing the number of sampled chunks reduces this distance, indicating better coverage of realized rollouts. Higher environmental stochasticity leads to larger distances, reflecting the increased difficulty of predicting future states under stronger execution noise.

{\bf\noindent Corrective behavior in demonstrations.}
We next study when \OURS is most effective. As \OURS selects among action chunks already sampled from the policy, its benefit depends on whether the policy distribution contains useful corrective behaviors and whether the auxiliary world model provides a latent space that can reliably distinguish the consequences of these candidate chunks. 
That is, 
\OURS requires the action-chunking policy to generate useful recovery options under stochastic dynamics. The Kinetix setting allows us to control this factor by training RL experts under different levels of action noise before collecting demonstrations. As shown in~\figref{fig:kinetix_expert_train_noise}, \OURS scales more effectively with the same number of sampled chunks when the imitation policy is trained on demonstrations from experts exposed to higher action noise. In contrast, when the policy is trained only on demonstrations from experts trained with low action noise, e.g., $\sigma=0.1$, increasing the sample count provides smaller gains. This suggests that test-time scaling is most useful when the demonstration data already contains corrective behaviors that the learned policy can sample at inference time.

{\bf\noindent Latent representation matters.}  We now study how the choice of auxiliary world model affects
\begin{wrapfigure}[15]{r}{0.5\textwidth}
  \centering
  \vspace{-10pt}
  \includegraphics[width=\linewidth]{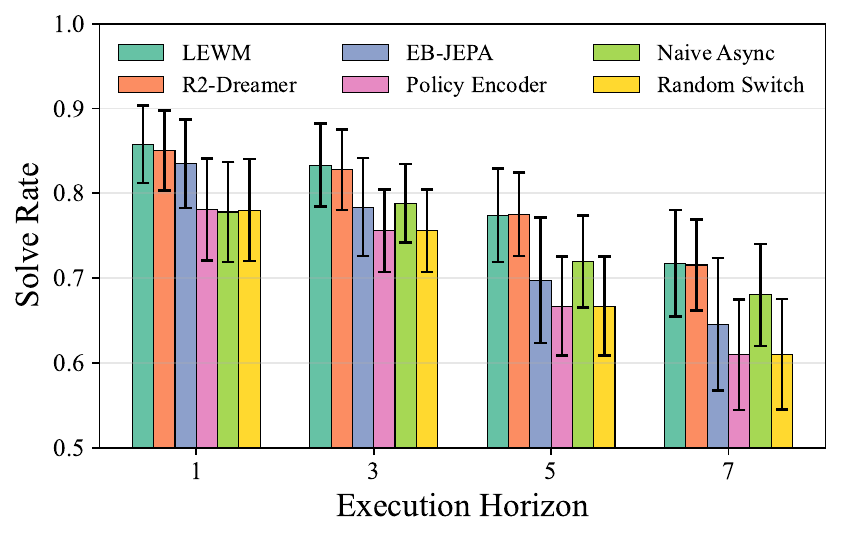}
  \vspace{-0.7cm}
  \caption{
    Ablations on different world model architectures. All methods are evaluated under action noise $\sigma=0.2$, $N=20$ for \OURS.
    }
  \label{fig:kinetix_wm_ablations}
  \vspace{-10pt}
\end{wrapfigure}
 \OURS. We evaluate  several decoder-free latent dynamics models, including an RSSM variant adapted from R2-Dreamer~\cite{morihira2026rdreamer},
LeWorldModel (LEWM)~\cite{maes2026leworldmodel}, and energy-based JEPA (EB-JEPA)~\cite{terver2026lightweight}. For R2-Dreamer, we remove the reward and continuation predictors and train with only the Barlow Twins, dynamics, and representation losses. For LEWM and EB-JEPA, we use one-step prediction with latent regularization: SigReg for LEWM and VICReg~\cite{bardes2022vicreg} for EB-JEPA. We also test a variant that freezes the action-chunking policy encoder and trains only a one-step latent dynamics predictor. As references, we include naive open-loop execution and a Random Switch baseline, which samples batched chunks but switches among them randomly during execution.

\figref{fig:kinetix_wm_ablations} shows that latent representation strongly affects \OURS performance. EB-JEPA degrades under longer execution horizons, likely because accumulated rollout errors make latent matching unreliable. Its performance remains above Random Switch, suggesting that its predictions retain some useful structure, but are insufficient for reliable long-horizon matching. In contrast, the frozen-policy-encoder variant regresses toward Random Switch, suggesting that these features, after the one-layer MLP projection, %
do not provide enough predictive information for identifying policy-intended future states. LEWM performs competitively and can even outperform the RSSM variant, indicating that \OURS does not require an explicitly stochastic world model. Its core requirement is a latent representation and dynamics model that supports reliable phase-aligned matching among candidate chunks.

\begin{figure}[t]
  \centering
  \includegraphics[width=1\textwidth]{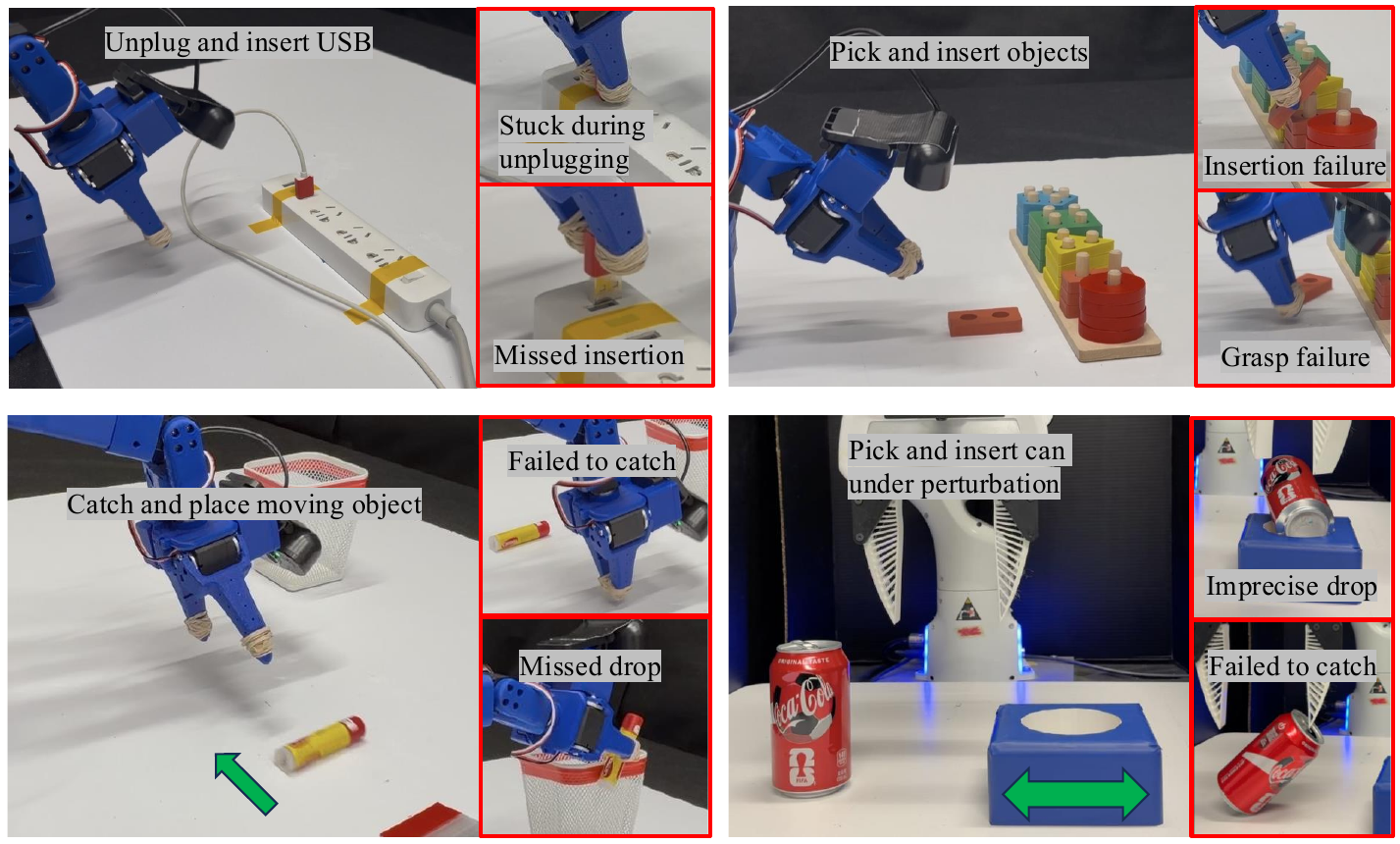}
  \caption{
    We design four hardware experiments on the SO-101 robot arm and Franka Emika Panda robot. The stochasticity mainly arises from inaccurate hardware execution on SO-101. In the moving-object grasping task, additional uncertainty comes from the uncertain velocity of the rolling object. In the last pick and insert can test, we manually perturb the target box before insertion. The figure shows representative failure modes caused by dynamic stochasticity. 
    }
  \label{fig:so101experiment}
  \vspace{-15pt}
\end{figure}

\subsection{Hardware Deployment}
\label{sec:hardware_experiment}

We evaluate \OURS on real robot hardware to study whether it improves action-chunking policies under stochastic dynamics in physical manipulation. We focus on three sources of uncertainty that commonly arise in hardware deployment: (i) noisy execution due to hardware limitations (we provide further discussion on servo precision in Appendix~\ref{app:hardware_uncertainty}), (ii) partial observability of task state, and (iii) external perturbations that alter the environment during execution. We design four manipulation tasks across the SO-101 and Franka robot arms, with representative failure modes shown in~\figref{fig:so101experiment}.
\vspace{-3pt}
\begin{itemize}[topsep=0pt, leftmargin=16pt]
    \setlength{\itemsep}{0.0pt}
    \setlength{\parskip}{2.5pt}

    \item [(1)] \textit{Unplug and insert USB (SO-101):} The robot unplugs a USB connector from the right port and inserts it into the left port. Limited motor precision can cause the connector to get stuck or miss the target port.

    \item [(2)] \textit{Catch and place moving object (SO-101):} The robot catches a rolling object and places it into a basket. Varying object velocity and single-frame observation make the dynamics partially observed, leading to missed catches or inaccurate drops.

    \item [(3)] \textit{Pick and insert object (SO-101):} The robot picks up a toy block and inserts it at the correct location. End-effector oscillations can cause grasp failures or inaccurate placement.

    \item [(4)] \textit{Insert can under perturbation (Franka):} The robot picks up a can and inserts it into a target slot while the container is manually perturbed. Failure to react can cause missed insertion, while imprecise grasping can lead to dropping the can.
\end{itemize}
{\bf Implementation details.}
We collect approximately 50 to 100 expert demonstration episodes per task through teleoperation. A key observation is consistent with our simulation results: \OURS is most effective when demonstrations contain corrective behaviors that can be sampled as recovery options under rollout deviations. We fine-tune SmolVLA~\cite{shukor2025smolvla} for the SO-101 experiments and $\pi_{0.5}$~\cite{black2025pi} for the Franka experiment. All hardware experiments use the R2-Dreamer latent world model formulation, trained on the same offline demonstration dataset used for policy training. During evaluation, we set the control rate to 30 Hz for SO-101 and 10 Hz for Franka, matching the frame rates used during demonstration collection. For the four tasks, we use execution horizons of 30, 20, 30, and 10 steps, respectively. These horizons are chosen to be long enough to avoid frequent idle transitions in the demonstrations, while remaining short enough to support timely inference in dynamic tasks.

{\bf Computation hardware.} We report the overhead of test-time scaling under two hardware setups. SO-101/SmolVLA latency was measured on a workstation with an NVIDIA RTX 5090 GPU and Intel Core Ultra 9 CPU. Franka/$\pi_{0.5}$ latency was measured in a distributed setup, with robot control on a local PC with an Intel i7-11700KF CPU and RTX 3080 GPU, and inference on a remote workstation with an AMD Threadripper PRO 5995WX CPU and RTX A6000 GPU communicating via WebSocket. All reported values are averaged over one test session.

{\bf Hardware results.}
\tabref{tab:hardware_success} reports the success rates on the hardware tasks using SmolVLA on the SO-101 platform, where \OURS uses $N=10$ candidate chunks. \OURS improves performance across all tasks, suggesting that reactive chunk selection helps mitigate uncertainty from hardware inaccuracy, partially observed object motion, and external perturbations. \begin{wraptable}[8]{r}{0.43\textwidth}
\centering
\small
\vspace{-0.35cm}
\caption{Success rates over 20 rollouts on three SO-101 tests with SmolVLA.}
\vspace{-0.15cm}
\label{tab:hardware_success}
\begin{tabular}{lcc}
\toprule
Task & Open-loop & Ours \\
\midrule
USB unplug/insert & 75\% & \textbf{95\%} \\
Pick moving object & 60\% & \textbf{80\%} \\
Pick and insert toy & 35\% & \textbf{45\%} \\
\bottomrule
\end{tabular}
\end{wraptable}
\tabref{tab:pi_performance_latency} further reports the performance and latency of $\pi_{0.5}$ on the Franka arm. Under local inference, \OURS with $N=5$ substantially improves the success rate over open-loop execution, although this setting uses sequential inference rather than parallel batched sampling and therefore introduces visible pauses during execution, as shown in the supplementary videos. 

Next, we also evaluate remote inference to enable parallel sampling with larger candidate pools. Despite the added communication delay (1s), remote \OURS still outperforms the local open-loop baseline. However, increasing $N$ further does not improve performance, likely because the added inference delay offsets the benefits of within-chunk reactivity. %
We provide experiment videos in the supplementary material and summarize the observed failure modes in \tabref{tab:hardware_failure_modes}.

\begin{table}[H]
\vspace{-0.3cm}
\begin{minipage}{0.54\textwidth}
\caption{Performance and inference latency for the can insertion task with $\pi_{0.5}$ on the Franka arm. %
}
\label{tab:pi_performance_latency}
\resizebox{1\linewidth}{!}{
\begin{tabular}{lccc}
\toprule
Method & $N$ & Inference Time & Performance \\
\midrule
Naive open loop (local) & 1 & 194.68 ms & 10\% \\
\OURS (local) & 5 & 634.77 ms & \textbf{65\%} \\
\OURS (remote) & 10 & 697.93 ms & 40\% \\
\OURS (remote) & 15 & 1001.27 ms & 30\% \\
\OURS (remote) & 20 & 1303.82 ms & 35\% \\
\bottomrule
\end{tabular}
}
\end{minipage}
\hfill
\begin{minipage}{0.43\textwidth}
\caption{Latency breakdown of \OURS's world model components.}
\label{tab:pi_wm_latency}
\resizebox{1\linewidth}{!}{
\begin{tabular}{lccc}
\toprule
$N$ & Encoder & Predictor & Latent Matching \\
\midrule
5  & 1.41 ms & 1.83 ms & 0.16 ms \\
10 & 1.40 ms & 2.74 ms & 0.17 ms \\
20 & 1.56 ms & 3.09 ms & 0.18 ms \\
\bottomrule
\end{tabular}
}
\end{minipage}
\vspace{-.3cm}
\end{table}

{\bf Inference latency.}
\tabref{tab:pi_performance_latency} also reports the inference latency of $\pi_{0.5}$ under different sample sizes and hardware settings. Policy inference becomes slower as the number of sampled chunks increases, but the measured latency does not grow linearly with $N$. In contrast, \tabref{tab:pi_wm_latency} shows that the world-model-related computation remains at the millisecond level: increasing the batched latent rollout from $N=5$ to $N=20$ adds only about $1.2$ ms of delay. We additionally report the inference latency of SmolVLA in~\tabref{tab:inference_latency}.

{\bf Nature of the corrective behaviors.}
Although we initially expected \OURS to produce more apparent within-chunk reactivity, such as switching to a substantially different behavior mode to catch a missed moving object as illustrated in Fig.~\ref{fig:hardware_demo}, the observed corrections are mostly local variations around the current action mode. This is consistent with the analysis by~\citet{pan2025much}, which suggests that the strength of generative policies often lies less in distributional learning and more in iterative refinement around likely actions. In our hardware experiments, sampled action chunks rarely exhibit distinct high-level modes. Instead, they mainly provide small corrective variations around an imperfect nominal behavior. Consequently, increasing the number of samples in \OURS helps primarily by exposing more of these local correction options, rather than by discovering qualitatively different strategies.

\section{Conclusion}
We presented \OURS, a test-time scaling method for improving the reactivity of action-chunking policies under stochastic dynamics. \OURS samples multiple candidate chunks, rolls out their latent consequences with a lightweight world model, and selects actions through latent matching at each step, enabling online correction. In Kinetix, our approach's gains increase with stochasticity and sample count, and depend on corrective behaviors in the demonstration data and reliable latent dynamics. On real hardware, \OURS improves performance across four manipulation tasks involving execution noise, partial observability, and external perturbations. These results suggest that lightweight predictive models can act as practical reactive modules for robust long-horizon VLA control.

\newpage
\bibliographystyle{ieee_fullname}  
\bibliography{references}

\newpage
\appendix
\setcounter{section}{0}
\renewcommand{\theHsection}{A\arabic{section}}
\renewcommand{\thesection}{A\arabic{section}}
\renewcommand{\thetable}{A\arabic{table}}
\setcounter{table}{0}
\setcounter{figure}{0}
\renewcommand{\thetable}{A\arabic{table}}
\renewcommand\thefigure{A\arabic{figure}}
\renewcommand{\theHtable}{A.Tab.\arabic{table}}%
\renewcommand{\theHfigure}{A.Abb.\arabic{figure}}%
\renewcommand\theequation{A\arabic{equation}}
\renewcommand{\theHequation}{A.Abb.\arabic{equation}}%

\section{Technical appendices and supplementary material}

\subsection{Additional Experiment Results}

\begin{figure}[h]
  \centering
  \includegraphics[width=\textwidth]{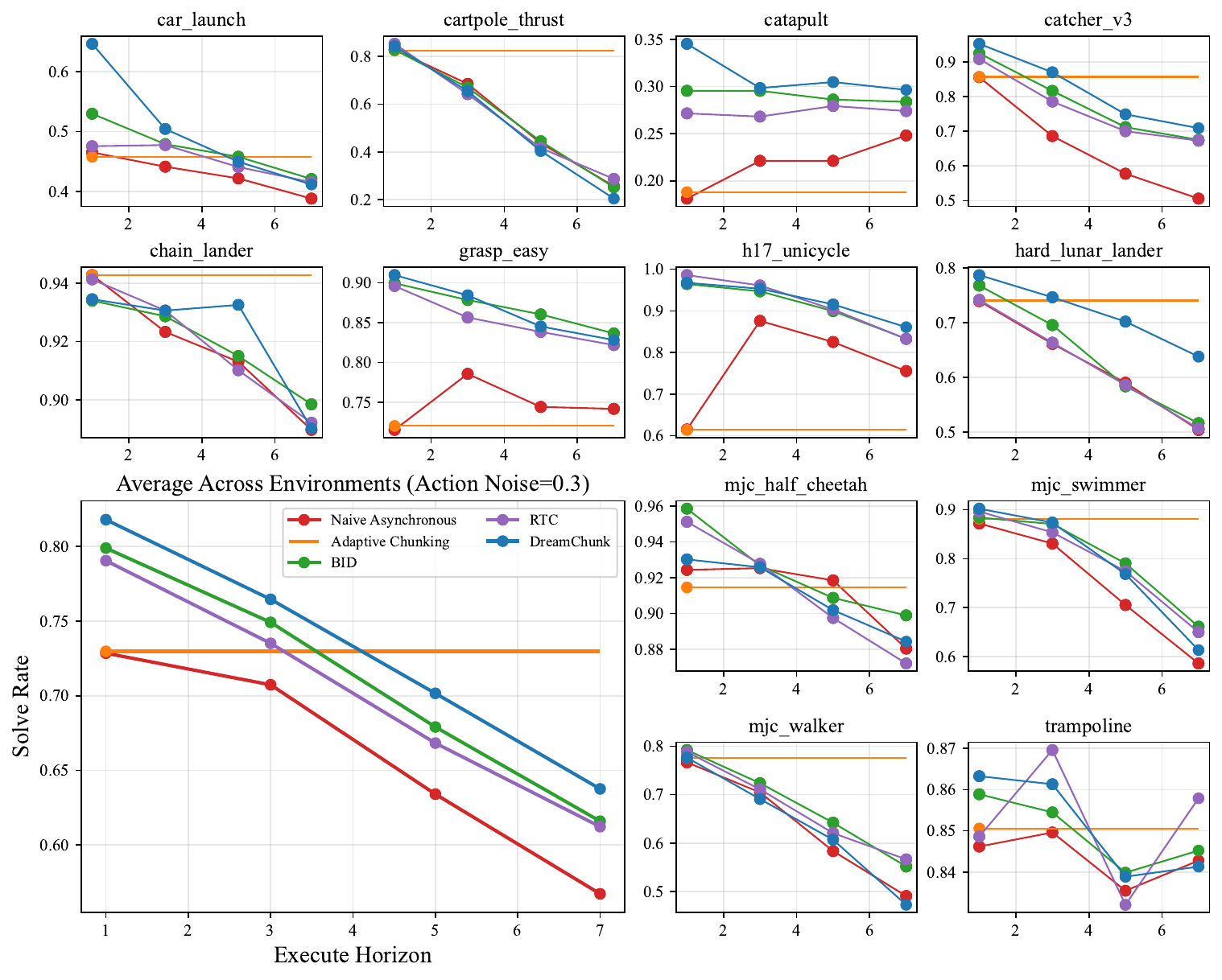}
  \caption{Solve rates across 12 environments in Kinetix with action noise $\sigma=0.2$. Adaptive chunking method is considered not to have a fixed execution horizon, and we report its performance as a horizontal line. }
  \label{fig:kinetix_executionhorizon}
\end{figure}

\begin{table}[h]
\centering
\captionsetup{width=0.85\textwidth}
\caption{Failure modes observed in hardware experiments.}
\label{tab:hardware_failure_modes}
\begin{tabular}{llcc}
\toprule
\textbf{Task} & \textbf{Failure mode} & \textbf{DreamChunk} & \textbf{Open-loop} \\
\midrule
\multirow{3}{*}{Toy stacking} 
& Inaccurate placement & 3 & 5 \\
& Missed grasp & 5 & 8 \\
& Stuck in placement & 3 & 0 \\
\midrule
\multirow{2}{*}{Grasp moving object} 
& Failed to intercept & 4 & 6 \\
& Inaccurate placement & 0 & 2 \\
\midrule
\multirow{2}{*}{USB unplug and insert} 
& Stuck at unplugging & 0 & 3 \\
& Incorrect-angle insertion & 1 & 2 \\
\midrule
\multirow{2}{*}{Can insertion (local $N=5$)} 
& Failed to pick up & 0 & 9 \\
& Missed insertion & 7 & 9 \\
\bottomrule
\end{tabular}
\end{table}

We provide the additional experiment results in this section. Figure.~\ref{fig:kinetix_executionhorizon} display the performance in 12 Kinetix environments at different execution horizon with inference delay $d=1$ and action noise $\sigma=0.2$.

As shown in Table~\ref{tab:inference_latency}, increasing the number of sampled chunks raises the action-inference latency of SmolVLA from 129.25 ms for the baseline to 303.70 ms, 326.72 ms, and 358.51 ms for \OURS with $N=5$, $10$, and $15$, respectively. In contrast, the additional world-model computation is lightweight: encoding, latent prediction, and latent matching take only 4.64 ms, 2.36 ms, and 0.20 ms, respectively.

\subsection{Implementation Details}
\begin{wraptable}{r}{0.43\textwidth}
\vspace{-11pt}
\centering
\small
\caption{Inference latency of SmolVLA and RSSM for SO-101 tests.}
\label{tab:inference_latency}
\begin{tabular}{lc}
\toprule
Component & Latency (ms) \\
\midrule
Baseline inference & 129.25 \\
Action inference, $N=5$ & 303.70 \\
Action inference, $N=10$ & 326.72 \\
Action inference, $N=15$ & 358.51 \\
\midrule
WM encoder & 4.64 \\
WM predictor & 2.36 \\
WM latent matching & 0.20 \\
\bottomrule
\end{tabular}
\vspace{-4pt}
\end{wraptable}
\paragraph{World model architecture in simulation.}
For the R2-Dreamer world model, we use MLP modules together with a GRU-like recurrent transition to process vector observation from Kinetix environment. Since the observation vector has length around 700, the observation encoder is a two-layer MLP with dimensions \(\texttt{obs\_dim} \rightarrow 512 \rightarrow 256\). The prior and posterior RSSM heads use hidden size 512, with two hidden layers for the prior and one hidden layer for the posterior by default. The deterministic dynamics are implemented with a blockwise GRU update using deterministic width 256, a stochastic state of size \(32 \times 16\) flattened to 512 dimensions, transition hidden size 512, one transition hidden layer, and 8 blocks. For deterministic JEPA-style world models including EB-JEPA and LEWM, we use a fully MLP-based architecture without recurrent dynamics. The model consists of an observation encoder \(\texttt{obs\_dim} \rightarrow 512 \rightarrow 256\), an action encoder \(\texttt{action\_dim} \rightarrow 256 \rightarrow 256\), a predictor with one hidden stage that maps the concatenated observation-action representation from \(256+256\) dimensions to a 512-dimensional hidden layer and then to a 256-dimensional output with residual connection and LayerNorm, and a projector \(256 \rightarrow 512 \rightarrow 256\). LEWM and EB-JEPA differ only in their latent regularization objective: LEWM uses SIGReg~\cite{maes2026leworldmodel}, whereas EB-JEPA uses variance-covariance regularization~\cite{terver2026lightweight}.

\paragraph{Hardware-induced execution uncertainty.}
\label{app:hardware_uncertainty}
We further clarify the uncertainty introduced by imprecise hardware execution in our experimental task design. The low-cost SO-101 arm uses Feetech STS3215 bus servos, whose manufacturer specifications report a nominal encoder resolution of $0.088^\circ$ and a gearbox backlash limit of $0.5^\circ$~\cite{huggingface2026so101}. An independent online test of the same servo model reported that practical backlash can exceed this nominal limit and lead to millimeter-scale end-effector displacement on an 86 mm single-link setup~\cite{robonine2025sts3215test}. Consistent with this observation, we found that the SO-101 end-effector position is affected by accumulated errors across multiple servo joints, making precise manipulation challenging even during teleoperation. As a point of comparison, the industrial-grade Franka Emika Panda provides substantially higher positioning precision: its official datasheet specifies a 7-DoF arm with 855 mm reach, 3 kg payload, and end-effector repeatability of $\pm 0.1$ mm~\cite{franka2018panda_datasheet}. This contrast highlights that our SO-101 experiments are conducted on a low-cost hardware platform with non-negligible actuation uncertainty.

\subsection{Analysis of Reactivity}
\label{app:reactivity_coverage}

\paragraph{Reactivity failure.}
We provide a simplified analysis to relate \OURS to the reactive objective in Eq.~(2). Recall that $J_{\mathrm{react}}(\eta)$ measures whether an execution strategy $\eta$ selects actions that remain aligned with the decision that the same policy would make if re-inferred from the realized observation. We say that a reactivity failure occurs at phase $\tau$ when the executed action has low alignment with the current policy decision,
\begin{equation}
    \mathrm{Align}(a^{\eta}_{t+\tau}, o_{t+\tau})
    <
    \epsilon_{\mathrm{align}},
\end{equation}
where $\epsilon_{\mathrm{align}}$ is an alignment threshold. This definition captures the central issue of open-loop chunk execution: stochastic dynamics, partial observability, or execution noise can move the system to a realized observation where the remaining actions in the originally sampled chunk no longer match what the policy would choose if queried again.

\paragraph{Open-loop misalignment.}
Let $p_{\mathrm{dyn}}$ denote the per-step probability that such a stochastic deviation causes the open-loop action from the original chunk to become misaligned with the policy decision conditioned on the realized observation. For a standard open-loop execution strategy $\eta_{\mathrm{open}}$ that executes the original chunk for horizon $H$, the probability that no such misalignment occurs during the execution window is
\begin{equation}
    P_{\mathrm{align}}^{\mathrm{open}}(H)
    =
    (1-p_{\mathrm{dyn}})^H .
\end{equation}
Therefore, the probability that open-loop execution encounters at least one phase that degrades $J_{\mathrm{react}}(\eta_{\mathrm{open}})$ is
\begin{equation}
    P_{\mathrm{misalign}}^{\mathrm{open}}(H)
    =
    1-(1-p_{\mathrm{dyn}})^H .
\end{equation}
This expression captures the benefit of shortening the execution horizon: a smaller $H$ reduces the chance that the policy continues executing stale actions before the next inference step.

\paragraph{Candidate coverage.}
\OURS takes a different approach. Instead of shortening the execution horizon, it keeps a longer horizon but samples multiple candidate chunks and uses the world model to maintain a pool of phase-aligned dreamed future states. Let $p_\delta$ denote the probability that one sampled chunk contains, at execution phase $\delta$, an action whose dreamed latent state is compatible with the realized observation and whose action is therefore aligned with the policy decision re-inferred at that observation. This probability is induced by the policy distribution and depends on whether the learned policy can sample corrective actions around perturbed rollouts. With $N$ independently sampled chunks, the probability that at least one candidate provides an aligned corrective action at phase $\delta$ is
\begin{equation}
    P_{\mathrm{cover}}(N,\delta)
    =
    1-(1-p_\delta)^N .
\end{equation}
This is the same coverage effect measured empirically by the nearest-dreamed-state distance: increasing $N$ makes it more likely that the dreamed-state pool contains a candidate close to the realized trajectory, which serves as a surrogate for higher alignment under $J_{\mathrm{react}}(\eta)$.

\paragraph{World-model reliability.}
However, coverage alone is not sufficient. \OURS also requires the auxiliary world model to provide a latent space in which phase-aligned matching can identify the candidate whose action is most compatible with the realized observation. We therefore introduce a reliability term $r_{\mathrm{wm}}\in[0,1]$, which represents the probability that latent matching selects an aligned corrective chunk when such a chunk exists in the candidate pool. The effective probability that a stochastic deviation can be corrected is approximated as
\begin{equation}
    P_{\mathrm{corr}}(N,\delta)
    =
    r_{\mathrm{wm}} P_{\mathrm{cover}}(N,\delta)
    =
    r_{\mathrm{wm}}\left[1-(1-p_\delta)^N\right].
\end{equation}
Accordingly, the per-step probability that a misalignment occurs and remains unrecovered is
\begin{equation}
    p_{\mathrm{unrec}}
    =
    p_{\mathrm{dyn}}\left(1-P_{\mathrm{corr}}(N,\delta)\right).
\end{equation}
For \OURS with a longer execution horizon $H_L$, the probability of encountering at least one unrecovered misalignment can be approximated as
\begin{equation}
    P_{\mathrm{fail}}^{\mathrm{DC}}(H_L,N)
    =
    1-
    \left(
    1-
    p_{\mathrm{dyn}}
    \left(
    1-r_{\mathrm{wm}}\left[1-(1-p_\delta)^N\right]
    \right)
    \right)^{H_L}.
\end{equation}

\paragraph{Comparison with shorter-horizon execution.}
Now consider a shorter-horizon baseline with execution horizon $H_S < H_L$. Since it does not maintain a candidate pool for within-horizon correction, its probability of encountering at least one misalignment is
\begin{equation}
    P_{\mathrm{fail}}^{\mathrm{short}}(H_S)
    =
    1-(1-p_{\mathrm{dyn}})^{H_S}.
\end{equation}
\OURS is expected to achieve higher reactivity than this shorter-horizon baseline when it has a smaller probability of unrecovered misalignment:
\begin{equation}
    P_{\mathrm{fail}}^{\mathrm{DC}}(H_L,N)
    <
    P_{\mathrm{fail}}^{\mathrm{short}}(H_S).
\end{equation}
Substituting the two failure probabilities gives
\begin{equation}
    1-
    \left(
    1-
    p_{\mathrm{dyn}}
    \left(
    1-r_{\mathrm{wm}}\left[1-(1-p_\delta)^N\right]
    \right)
    \right)^{H_L}
    <
    1-(1-p_{\mathrm{dyn}})^{H_S}.
\end{equation}
Equivalently,
\begin{equation}
    \left(
    1-
    p_{\mathrm{dyn}}
    \left(
    1-r_{\mathrm{wm}}\left[1-(1-p_\delta)^N\right]
    \right)
    \right)^{H_L}
    >
    (1-p_{\mathrm{dyn}})^{H_S}.
\end{equation}

For small $p_{\mathrm{dyn}}$, using the first-order approximation $(1-x)^H \approx 1-Hx$, this condition becomes
\begin{equation}
    H_L p_{\mathrm{dyn}}
    \left(
    1-r_{\mathrm{wm}}\left[1-(1-p_\delta)^N\right]
    \right)
    <
    H_S p_{\mathrm{dyn}}.
\end{equation}
When $p_{\mathrm{dyn}}>0$, we can cancel $p_{\mathrm{dyn}}$ and obtain
\begin{equation}
    r_{\mathrm{wm}}\left[1-(1-p_\delta)^N\right]
    >
    1-\frac{H_S}{H_L}.
    \label{eq:dreamchunk_horizon_condition}
\end{equation}
This inequality gives a simple interpretation of the trade-off. A longer-horizon \OURS strategy can compensate for its larger exposure to stochastic deviations if its candidate pool is sufficiently likely to contain an aligned corrective action and if the world model can reliably identify that candidate. In other words, shortening the execution horizon reduces the probability that stale actions become misaligned, whereas \OURS increases the probability that an aligned correction is available when misalignment occurs.

\paragraph{Implications.}
This analysis also clarifies when \OURS should be effective. The coverage term $P_{\mathrm{cover}}(N,\delta)$ depends on the diversity of sampled action chunks and on whether the demonstration data contains corrective behaviors that the learned policy can sample. If the policy distribution only produces small variations around a nominal trajectory and the demonstrations do not contain recovery behaviors, then $p_\delta$ remains small and increasing $N$ provides limited benefit. In contrast, when demonstrations include reactions to stochastic dynamics, increasing $N$ can expose more local correction options and improve the chance of maintaining high $J_{\mathrm{react}}(\eta)$. The world-model reliability term $r_{\mathrm{wm}}$ is equally important: even if the candidate pool contains an aligned corrective action, \OURS can only use it when the learned latent dynamics and matching metric correctly identify the candidate whose dreamed state is compatible with the realized observation. Meanwhile, $p_{\mathrm{dyn}}$ is determined by the environment and hardware: stronger action noise, partial observability, external perturbations, or low-precision actuation all increase the probability that open-loop actions become misaligned with the realized observation. Thus, \OURS is most useful in an intermediate regime where stochastic deviations occur often enough that open-loop execution becomes brittle, but the learned policy and world model still provide enough corrective coverage and matching reliability for latent matching to recover.

\subsection{ Discussion}
{\bf\noindent Limitation.} The proposed method requires additional computational resources at test time to sample batched action chunks, which is deployable using remote reasoning on a PC but the sample size would be much more limited at edge device. We also discussed the nature of corrective behavior is largely confined by the expressiveness of generative policies, our experiment shows the sampled action chunks does not have distinct mode different, limiting \OURS to demonstrate significant strategy change during within-chunk execution. Further, the effectiveness of proposed methods depends on whether expert demonstration includes responses to stochastic dynamics, posing requirement in data collection.

{\bf\noindent Extension.}
We aim to let the proposed method innovate the use of world models in VLA studies, taken inspirations from well-established model-based robotics studies like model predictive control, robust control, and model predictive path integral~\cite{alvarez2025real}. Since \OURS method samples multiple action chunks at test-time, it can also be integrated with other test-time scaling algorithms. For instance, adding the backward loss from BID ~\cite{liu2025bidirectional} into chunk selection can improve cross-chunk consistency, apply cross-chunk entropy~\cite{liang2026adaptive}, and provide measures on adaptively changing execution horizon. To reduce the overhead of batched action sampling, future implementations could amortize VLM backbone inference by computing the visual-language embedding once and reusing it across candidate chunks, while sampling different noise inputs in the action expert to generate diverse actions.

{\bf\noindent Social impact.}
This work contributes to embodied intelligence and robot learning by improving the robustness of robot policies under stochastic dynamics, hardware uncertainty, and partial observability. Such capabilities could benefit assistive robotics, manufacturing, and logistics, where safer and more adaptive robot behavior is valuable. However, more robust robot policies may also increase risks in harmful or poorly regulated applications, including weaponized automation, unsafe autonomous systems. Broader advances in robotic automation may also affect labor demand in some sectors. These risks are not unique to \OURS, but motivate task-specific safety constraints, human oversight, misuse monitoring, and deployment restrictions in high-risk domains.

\newpage

\newpage

\end{document}